\documentclass{article}

\usepackage{arxiv}

\usepackage[utf8]{inputenc} % allow utf-8 input
\usepackage[T1]{fontenc}    % use 8-bit T1 fonts
\usepackage{hyperref}       % hyperlinks
\usepackage{url}            % simple URL typesetting
\usepackage{booktabs}       % professional-quality tables
\usepackage{amsfonts}       % blackboard math symbols
\usepackage{nicefrac}       % compact symbols for 1/2, etc.
\usepackage{microtype}      % microtypography
\usepackage{lipsum}
\usepackage{graphicx}
\graphicspath{ {./images/} }

\title{A genetic algorithm for student academic resource allocation}

\author{
 Ana F. Hernández \\
  Dept. Developmental and Educational Psychology\\
  University of Murcia, Spain\\
  \texttt{directora@aventurinna.com} \\
   \And
 Andrej Franulic \\
  Comenius iDi SRL, Spain\\
  \texttt{it@aventurinna.com} \\
  \And
 Fernando Jiménez \\
  Dept. Information and Communications Engineering\\
  University of Murcia, Spain \\
  \texttt{fernan@um.es} \\
}
\usepackage{xcolor}
\usepackage{graphicx}
\usepackage{amsmath}
\usepackage{amssymb}
\usepackage{mathtools}
\usepackage{algorithm}
\usepackage{algpseudocode}

\begin{document}
\maketitle
\begin{abstract}
The optimal allocation of academic resources to individual students is essential for addressing learner diversity and fostering equitable educational outcomes. Within the framework of the Erasmus+ KA220-SCH project, this paper models the selection of educational materials for high school mathematics students as a 0--1 binary combinatorial optimization problem subject to strict study time constraints. Given the NP-hard complexity of the formulation, exact solution methods become computationally intractable as resource catalogs scale. To address this challenge, we propose a Genetic Algorithm integrated with a specialized constraint repair mechanism to effectively search the binary decision space. Experimental evaluation across 10 independent runs demonstrates fast convergence, high solution quality, and strong algorithmic stability across different base seeds. These results confirm the practical utility of metaheuristic approaches for real-time decision-support systems in secondary education.
\end{abstract}

% keywords can be removed

\section{Introduction}

Educational frameworks increasingly demand personalized learning pathways to accommodate student diversity and foster equitable access to quality education. Within the scope of the \textbf{Erasmus+ KA220-SCH — Cooperation Partnerships in School Education} project, this work addresses the challenge of optimizing educational resource allocation problems to enhance learning outcomes \cite{5524021, 10220065}, specifically for high school mathematics students. The project promotes equity and diversity by addressing the needs of high-achieving students, those with learning difficulties, and those with special educational needs. Additionally, it seeks to support STEM educators by developing a resource bank for lesson planning and strategies to reduce student exam anxiety. The initiative targets secondary schools in the Region of Murcia (Spain), as well as partner institutions in Italy and Portugal.

From a computational perspective, matching a specific student profile with an ideal set of educational materials can be modeled as a \textbf{0--1 binary combinatorial optimization problem} \cite{Kellerer2004}. Given a predefined catalog of pedagogical resources categorized by teachers, the goal is to select an optimal subset of resources that maximizes educational benefit while satisfying operational constraints (such as maximum available study time). 

Formally, for a catalog containing $n$ available resources, the decision space corresponds to binary vectors $\mathbf{x} \in \{0, 1\}^n$, yielding a search space of $2^n$ possible candidate solutions. Due to its structural equivalence to classic knapsack-type decision models, the problem is \textbf{NP-hard} \cite{Garey1979}. As $n$ increases, exact enumerative or deterministic methods become computationally intractable for real-time or large-scale decision support systems.

To efficiently explore this exponentially large search space, metaheuristic optimization techniques are required. \textbf{Genetic Algorithms} (GAs) \cite{Goldberg1989, Holland1992}, which are population-based stochastic search algorithms inspired by the principles of natural selection and genetics, are particularly well-suited for such binary combinatorial landscapes.
By maintaining a population of candidate solutions (chromosomes) and evolving them through iterative operators such as selection, crossover, and mutation, GAs balance global exploration of the search space with local exploitation of promising regions without requiring gradient information.

The remainder of this technical report is structured as follows:
\begin{itemize}
	\item \textbf{Section 2} provides the formal mathematical formulation of the optimization problem, detailing the objective function and operational constraints.
	\item \textbf{Section 3} describes the components of the proposed GA, including chromosome representation, genetic operators, and the specialized constraint repair mechanism.
	\item \textbf{Section 4} outlines the implementation details and presents the experimental evaluation conducted on synthetically generated resource catalogs and student profiles.
	\item \textbf{Section 5} summarizes the conclusions and highlights potential avenues for future work.
\end{itemize}

\section{Problem definition}

We consider a single-student academic resource allocation problem. The objective is to select an optimal subset of academic resources from a pre-established finite catalog to support and personalize the student's learning process in mathematics. The optimization model is executed independently for each individual student.

\subsection{Sets and indices}

The foundational sets and discrete spaces defining the system domain are formalized as follows:
\begin{itemize}
	\item $\mathcal{R} = \{1,\dots,n\}$: set of available candidate academic resources.
	\item $\mathcal{A} = \{\text{Numbers}, \text{Algebra}, \text{Geometry}, \text{Statistics}, \text{Functions}\}$: set of fundamental mathematical areas.
	\item $\mathcal{C} = \{1,2,3,4\}$: set of secondary education grade levels (courses).
	\item $\mathcal{L} = \{1,2,3\}$: set of academic proficiency levels, corresponding to 
	$1 =$ Basic, 
	$2 =$ Full, and 
	$3 =$ High.
	\item $\mathcal{T} = \{\text{Theory}, \text{Practice}\}$: set of pedagogical resource categories.
\end{itemize}

Throughout the model, the indexing conventions are defined as:
\[
i \in \mathcal{R}, \quad
j \in \mathcal{A}, \quad
\text{and} \quad
\ell \in \mathcal{L}.
\]

\subsection{Student data parameters}

The student profile is characterized by academic performance metrics and pedagogical diagnostics.

\subsubsection{Academic profile}

A student is represented by the tuple of parameters:
\begin{itemize}
	\item $c \in \mathcal{C}$: current educational grade level of the student.
	\item $GE_j \in \{1,\dots,15\}$: global evaluation score obtained by the student in mathematical area $j \in \mathcal{A}$.
	\item $L_j \in \mathcal{L}$: discrete proficiency level of the student in area $j \in \mathcal{A}$, derived directly from $GE_j$ according to the mapping:
\end{itemize}

\begin{equation}
	L_j =
	\begin{cases}
		1, & \text{if } 1 \le GE_j < 5, \\
		2, & \text{if } 5 \le GE_j < 10, \\
		3, & \text{if } 10 \le GE_j \le 15.
	\end{cases}
\end{equation}

\subsubsection{Pedagogical diagnosis}

To model specialized learning needs, we incorporate an expert-defined diagnostic factor $s \in (0,1]$. This scalar parameter reflects the student's learning condition and modulates the impact of pedagogical adjustments. Specifically, $s=1.0$ is assigned to high-ability students (no penalty applied), while values $s < 1.0$ represent increasing levels of learning difficulty or cognitive constraints. The diagnostic factors utilized in this framework are summarized in Table~\ref{tab:diagnoses}.

\begin{table}[!ht]
	\centering
	\begin{tabular}{lcl}
		\hline
		\textbf{Diagnosis} & \textbf{$s$} & \textbf{Pedagogical Rationale} \\
		\hline
		High Capabilities & 1.0 & High adaptability; minimal restriction across resource types. \\
		Regular Student & 0.8 & Standard cognitive profile without specific learning difficulties. \\
		Dyslexia & 0.7 & Symbolic/algebraic reading barriers; offset by visual-spatial strengths. \\
		ADHD & 0.6 & Inattention risks; requires concise, highly structured sequential tasks. \\
		Dyscalculia & 0.5 & Severe impediment in numerical processing and basic arithmetic. \\
		\hline
	\end{tabular}
	\caption{Diagnostic factors ($s$) and rationale based on student learning profiles.}
	\label{tab:diagnoses}
\end{table}

\subsection{Resource catalog properties and pre-filtering}

Each candidate resource $i \in \mathcal{R}$ in the master catalog is characterized by a tuple of static attributes:
\begin{itemize}
	\item $c_i \in \mathcal{C}$: target educational grade level of resource $i$.
	\item $a_i \in \mathcal{A}$: primary mathematical subject area of resource $i$.
	\item $\ell_i \in \mathcal{L}$: difficulty level associated with resource $i$.
	\item $t_i \in \mathcal{T}$: pedagogical category (Theory or Practice).
	\item $\tau_i \in \mathbb{N}_{+}$: estimated execution time in minutes, bounded by $t_{\min} \le \tau_i \le t_{\max}$.
\end{itemize}

Prior to the optimization run, a deterministic pre-filtering phase trims the master catalog $\mathcal{R}$ to retain only pedagogically suitable items for the student's current standing $c$ and area proficiency $L_{a_i}$. Specifically, a resource $i$ is retained in the decision space if and only if:
\begin{equation}
	\ell_i \ge L_{a_i} \quad \text{and} \quad c \le c_i \le c+1, \qquad \forall i \in \mathcal{R}.
	\label{eq:prefiltering}
\end{equation}

\subsection{Decision variables}

The selection process is governed by binary selection decision variables:
\begin{equation}
	x_i =
	\begin{cases}
		1, & \text{if resource } i \text{ is selected for the student}, \\
		0, & \text{otherwise},
	\end{cases}
	\quad \forall i \in \mathcal{R}.
\end{equation}

Consequently, a full candidate solution is represented by the binary decision vector:
\begin{equation}
	\mathbf{x} = (x_1, x_2, \dots, x_n) \in \{0,1\}^n.
\end{equation}

\subsection{Constraints}

To ensure that the recommended resource package is practically feasible for the student, we impose a global operational time budget. The cumulative estimated completion time of all selected resources must not exceed the maximum allowed duration $T_{\max}$:
\begin{equation}
	g_1(\mathbf{x}) = \sum_{i \in \mathcal{R}} \tau_i x_i - T_{\max} \le 0.
	\label{eq:constraint_time}
\end{equation}

\subsection{Objective function}

The objective function $f(\mathbf{x})$ quantifies the total educational utility gained by assigning a solution vector $\mathbf{x}$. It is designed to maximize pedagogical efficiency by prioritizing weak subject areas while penalizing excessive gaps in difficulty or grade alignment:

\begin{equation}
    f(\mathbf{x}) = \sum_{j \in \mathcal{A}} \frac{1}{GE_j}
    \sum_{\substack{i \in \mathcal{R} \\ a_i = j}}
    \frac{s}{s + (c_i - c) + (\ell_i - L_j)} x_i.
    \label{eq:objective-function}
\end{equation}

The structural design of Equation~\ref{eq:objective-function} operates through the following mechanisms:
\begin{itemize}
    \item \textbf{Area Need Weighting ($\frac{1}{GE_j}$):} Inversely scales utility with respect to the student's global evaluation score in area $j$. Deficits in basic competencies ($GE_j \approx 1$) yield higher rewards upon resource assignment compared to mastered areas ($GE_j \approx 15$).

    \item \textbf{Grade Level Alignment ($c_i-c$):} Favors resources matching the student's current grade ($c_i=c$) over resources targeting the following grade ($c_i=c+1$). The grade mismatch contributes to the denominator of the resource utility weight, reducing its value as the mismatch increases.

    \item \textbf{Difficulty Progression ($\ell_i-L_j$):} Rewards resources aligned with the student's proficiency level ($\ell_i=L_j$), while penalizing resources requiring higher proficiency levels ($\ell_i>L_j$). This difficulty mismatch contributes to the same denominator as the grade mismatch, jointly determining the resource utility weight.

    \item \textbf{Diagnostic Adjustment ($s$):} The diagnostic factor $s \in (0,1]$ controls the penalization of resources whose grade or difficulty level deviates from the student's academic profile. For perfectly aligned resources, the corresponding weight is $1$, independently of $s$. For resources with positive mismatches, lower values of $s$ increase the penalization. Consequently, $s$ modifies the relative utilities of resources rather than uniformly scaling them, potentially leading to different optimal allocations for different diagnostic profiles.
\end{itemize}

\subsection{Complete mathematical model}

Synthesizing the objective function and operational constraints, the single-student resource allocation problem is formally expressed as the binary 0--1 integer linear program:

\begin{equation}
    \begin{aligned}
        \max_{\mathbf{x}} \quad & f(\mathbf{x}) =
        \sum_{j \in \mathcal{A}} \frac{1}{GE_j}
        \sum_{\substack{i \in \mathcal{R} \\ a_i = j}}
        \frac{s}{s + (c_i - c) + (\ell_i - L_j)} x_i \\
        \text{s.t.} \quad &
        \sum_{i \in \mathcal{R}} \tau_i x_i \le T_{\max}, \\
        & x_i \in \{0,1\}, \quad \forall i \in \mathcal{R}.
    \end{aligned}
    \label{eq:final-problem}
\end{equation}

\section{Genetic algorithm design}

To solve the binary combinatorial optimization problem formulated in Section~\ref{eq:final-problem}, we implement a tailored GA. The algorithm balances exploration of the exponential search space $2^n$ with local exploitation of high-utility resource combinations. This section details the candidate representation, constraint handling mechanism, fitness mapping, mating selection, genetic operators, and environmental selection, whose global execution workflow is summarized in Algorithm~\ref{alg:ga}.

\begin{algorithm}[!ht]
	\caption{Genetic Algorithm for Academic Resource Allocation}\label{alg:ga}
	\begin{algorithmic}[1]
		\Require Problem domain data $(c, GE, L, s, \mathcal{R}, T_{\max})$, where $\mathcal{R}$ is the filtered resource catalog compatible with the student profile;
		\Require Population size $N$, Number of generations $G$, Crossover probability $p_c$, Mutation probability $p_m$.
		\Ensure Optimal binary resource allocation vector $\mathbf{x}^*$ and best utility $f^*$.
		\State $n \leftarrow |\mathcal{R}|$ \Comment{Search space dimension}
		\State $P_0 \leftarrow \text{InitializePopulation}(N, n)$ \Comment{Uniform binary initialization}
		\State $P_0 \leftarrow \text{RepairMaxTime}(P_0, \boldsymbol{\tau}, T_{\max})$ \Comment{Enforce initial feasibility}
		\State EvaluateFitness($P_0, c, GE, L, s, \mathcal{R}$)
		\State $t \leftarrow 0$
		\While{$t < G$}
		\State $M_t \leftarrow \text{BinaryTournamentSelection}(P_t, N)$
		\State $Q_t \leftarrow \text{TwoPointCrossover}(M_t, p_c)$
		%\State $Q_t \leftarrow \text{RepairMaxTime}(Q_t, \boldsymbol{\tau}, T_{\max})$ \Comment{Repair crossover offspring}
		\State $Q_t \leftarrow \text{BitflipMutation}(Q_t, p_m)$
		\State $Q_t \leftarrow \text{RepairMaxTime}(Q_t, \boldsymbol{\tau}, T_{\max})$ \Comment{Repair mutated offspring}
		\State EvaluateFitness($Q_t, c, GE, L, s, \mathcal{R}$)
		\State $P_{t+1} \leftarrow \text{SelectTopK}(P_t \cup Q_t, N)$ \Comment{$(\mu + \lambda)$ elitist replacement}
		\State $t \leftarrow t + 1$
		\EndWhile
		\State $\mathbf{x}^* \leftarrow \arg\max_{\mathbf{x} \in P_{G_{\max}}} \text{Fitness}(\mathbf{x})$
		\State \Return $\mathbf{x}^*$, $f(\mathbf{x}^*)$
	\end{algorithmic}
\end{algorithm}

\subsection{Chromosome representation and population initialization}

A candidate solution (individual or chromosome) is represented directly as a $n$-dimensional binary vector $\mathbf{x} = (x_1, x_2, \dots, x_n) \in \{0,1\}^n$, where $n = |\mathcal{R}|$ denotes the total number of items in the pre-filtered resource catalog. A gene value $x_i = 1$ indicates that resource $i$ is assigned to the student's learning plan, whereas $x_i = 0$ signifies non-selection.

The initial population $P_0$ of size $N = \text{\texttt{pop\_size}}$ is generated uniformly at random:
\begin{equation}
	x_i \sim \text{Bernoulli}(0.5), \quad \forall i \in \{1, \dots, n\}, \quad \forall \mathbf{x} \in P_0.
\end{equation}

\subsection{Repair mechanism for constraint handling}

Candidate solutions generated during uniform initialization or through variation operators (crossover and mutation) may exceed the maximum allowed study time budget $T_{\max}$. To maintain feasibility without relying on unguided fitness penalties, constraint handling is strictly enforced via a specialized repair operator, RepairMaxTime, formalized in Algorithm~\ref{alg:repair}.

\begin{algorithm}[!ht]
	\caption{Constraint Repair Operator (RepairMaxTime)}\label{alg:repair}
	\begin{algorithmic}[1]
		\Require Population/Chromosome binary matrix $\mathbf{X}$, Resource completion times $\boldsymbol{\tau} = (\tau_1, \dots, \tau_n)$, Max time limit $T_{\max}$.
		\Ensure Feasible population/chromosome binary matrix $\mathbf{X}'$ satisfying $\sum_{i=1}^n \tau_i x'_i \le T_{\max}$.
		\For{each individual $\mathbf{x}$ in $\mathbf{X}$}
		\State $T_{\text{total}} \leftarrow \sum_{i=1}^n \tau_i \cdot x_i$ \Comment{Compute total time for current solution}
		\If{$T_{\text{total}} > T_{\max}$}
		\State $I_{\text{active}} \leftarrow \{ i \in \{1, \dots, n\} \mid x_i = 1 \}$ \Comment{Extract active resource indices}
		\State $\text{RandomShuffle}(I_{\text{active}})$ \Comment{Randomize removal sequence}
		\State $idx \leftarrow 1$
		\While{$T_{\text{total}} > T_{\max}$ \textbf{and} $idx \le |I_{\text{active}}|$}
		\State $j \leftarrow I_{\text{active}}[idx]$
		\State $x_j \leftarrow 0$ \Comment{Deactivate resource}
		\State $T_{\text{total}} \leftarrow T_{\text{total}} - \tau_j$
		\State $idx \leftarrow idx + 1$
		\EndWhile
		\EndIf
		\EndFor
		\State \Return $\mathbf{X}$
	\end{algorithmic}
\end{algorithm}

The repair procedure operates directly on binary decision vectors. When a solution violates the time constraint ($\sum_{i=1}^n \tau_i x_i > T_{\max}$), the operator randomly deactivates selected resources ($x_i \leftarrow 0$) without replacement until the cumulative duration strictly complies with $T_{\max}$.

To guarantee that only feasible solutions enter the evaluation phase and progress through the evolutionary cycle, this repair operator is systematically applied at two strategic stages:
\begin{enumerate}
	\item \textbf{Post-Initialization:} Immediately following the uniform random generation of the initial population $P_0$.
	\item \textbf{Post-Mutation:} Immediately after offspring solutions undergo bit-flip mutation, correcting any newly activated genes ($0 \to 1$) that may have caused a budget overflow.
\end{enumerate}

\subsection{Fitness function}
Because the constraint repair mechanism guarantees that any chromosome evaluated by the algorithm satisfies $g_1(\mathbf{x}) \le 0$, the fitness function directly maps a repaired binary solution $\mathbf{x}$ to its corresponding objective function value $f(\mathbf{x})$.  This direct evaluation mechanism avoids the need for external penalty factors, ensuring that the selection operators operate exclusively on true educational utility values.

\subsection{Evolutionary mechanisms and operators}

The evolutionary cycle combines variation operators tailored to binary search spaces with selection mechanisms to drive convergence:

\begin{itemize}
	\item \textbf{Mating Selection (Binary Tournament):} Parental selection is conducted via Binary Stochastic Tournament Selection~\cite{Blickle1996}. Pairs of individuals are sampled uniformly at random from the population, and the one exhibiting the higher fitness value is selected to enter the mating pool.
	
	\item \textbf{Crossover (Two-Point Crossover):} Recombination is executed using Two-Point Crossover \cite{davis1991} with probability $p_c$. Two crossover points $k_1, k_2 \in \{1, \dots, n-1\}$ ($k_1 < k_2$) are selected uniformly at random, swapping the segment between $k_1$ and $k_2$ between the parent binary vectors to yield two offspring.
	
	\item \textbf{Mutation (Bit-Flip Mutation):} Variational noise is injected via Bit-Flip Mutation ~\cite{davis1991} with a per-gene mutation probability $p_m$. Each bit $x_i$ in an offspring chromosome is independently inverted ($x_i \leftarrow 1 - x_i$) with a per-gene mutation probability $p_m = 1/n$, where $n$ is the number of available resources.
	
	\item \textbf{Environmental Selection ($(\mu + \lambda)$-Survival):} Replacement follows an elitist $(\mu + \lambda)$ Evolution Strategy framework \cite{Back2000}, where $\mu = \lambda = N = \text{\texttt{pop\_size}}$. In each generation $t$, the current parent population $P_t$ ($\mu$) and the generated offspring population $Q_t$ ($\lambda$) are merged ($P_t \cup Q_t$). The top $N$ fittest individuals from the combined pool of size $2N$ are selected to form $P_{t+1}$, guaranteeing monotonic fitness convergence.
\end{itemize}

\section{Experimental evaluation and results}

The proposed optimization approach was implemented using the \texttt{pymoo} framework \cite{9078759}, specifically employing the GA module with duplicate elimination enabled. To evaluate the performance and convergence capabilities of the proposed GA, a systematic experimental study was conducted. This section details the experimental setup, including the student profile, catalog parameters, algorithm hyperparameters, and numerical findings obtained across multiple independent evolutionary runs. 

\subsection{Experimental setup and parameter configuration}

A synthetic master catalog containing $n = 1000$ candidate mathematical resources was generated, with estimated completion times $\tau_i$ uniformly bounded between $t_{\min} = 60$ minutes and $t_{\max} = 240$ minutes. 

The optimization was evaluated on a target student profile suffering from dyslexia. Table~\ref{tab:student_profile} details the specific academic indicators and diagnostic settings used for this experiment.

\begin{table}[h!]
	\centering
	\resizebox{\columnwidth}{!}{% <--- Cambiado a \columnwidth
		\begin{tabular}{ccccccccc}
			\hline
			\textbf{Student ID} & \textbf{Course ($c$)} & \textbf{Diagnosis} & \textbf{$s$} & \textbf{$GE_{\text{Numb}}$} & \textbf{$GE_{\text{Alg}}$} & \textbf{$GE_{\text{Geom}}$} & \textbf{$GE_{\text{Stat}}$} & \textbf{$GE_{\text{Funct}}$} \\
			\hline
			1 & 2 & Dyslexia & 0.7 & 12 & 13 & 9 & 10 & 12 \\
			\hline
		\end{tabular}%
	}
	\caption{Academic profile and diagnostic parameters for Student ID 1.}
	\label{tab:student_profile}
\end{table}

Prior to executing the evolutionary search, the master catalog was processed using the pre-filtering criteria defined in Equation~\ref{eq:prefiltering}. Out of the initial $n = 1000$ items, exactly $|\mathcal{R}| = 217$ resources met the grade level and proficiency requirements for Student 1, forming the active decision space ($n = 217$). The available time budget for the student was set to $T_{\max} = 8100$ minutes (135 hours). The GA was configured using the hyperparameters summarized in Table~\ref{tab:ga_params}.

\begin{table}[h!]
	\centering
	\begin{tabular}{lc}
		\hline
		\textbf{Parameter} & \textbf{Value} \\
		\hline
		Population size ($N$) & 50 \\
		Number of generations ($G$) & 300 \\
		Crossover probability ($p_c$) & 0.8 \\
		Mutation probability ($p_m$) & $1/n$ \\
		Number of runs ($K$) & 10 \\
		Base random seed & 1 (Incremented by 10 per run) \\
		\hline
	\end{tabular}
	\caption{Hyperparameter settings for the Genetic Algorithm.}
	\label{tab:ga_params}
\end{table}

To account for the stochastic nature of metaheuristics, $K = 10$ independent runs were executed, each initialized with a different random seed to ensure statistical diversity.

\subsection{Numerical results and statistical summary}

The objective values ($f(\mathbf{x})$) obtained across all 10 independent executions are detailed in Table~\ref{tab:runs_results}. Table~\ref{tab:statistical_summary} summarizes the key performance indicators across the experimental batch. The low standard deviation ($\sigma = 0.013338$) highlights the high consistency of the algorithm across varying random initializations.

\begin{table}[h!]
    \centering
    \begin{tabular}{ccc}
        \hline
        \textbf{Run} & \textbf{Seed} & \textbf{Best Fitness $f(\mathbf{x})$} \\
        \hline
        1  & 1  & 6.0286 \\
        2  & 11 & 6.0509 \\
        3  & 21 & 6.0687 \\
        4  & 31 & \textbf{6.0753} \\
        5  & 41 & 6.0413 \\
        6  & 51 & 6.0531 \\
        7  & 61 & 6.0676 \\
        8  & 71 & 6.0625 \\
        9  & 81 & 6.0644 \\
        10 & 91 & 6.0544 \\
        \hline
    \end{tabular}
    \caption{Best fitness values achieved across 10 independent algorithmic runs.}
    \label{tab:runs_results}
\end{table}

\begin{table}[h!]
    \centering
    \begin{tabular}{lc}
        \hline
        \textbf{Metric} & \textbf{Value} \\
        \hline
        Best Fitness ($f_{\text{best}}$) & 6.075289 \\
        Worst Fitness ($f_{\text{worst}}$) & 6.028632 \\
        Average Fitness ($f_{\text{avg}}$) & 6.056687 \\
        Standard Deviation ($\sigma$) & 0.013338 \\
        \hline
    \end{tabular}
    \caption{Statistical summary of fitness values over 10 independent runs.}
    \label{tab:statistical_summary}
\end{table}

\subsection{Analysis of the best solution}

The best solution was achieved during \textbf{Run 4} (Seed 31), attaining a peak utility score of $f(\mathbf{x}^*) = 6.075289$.
Out of the 217 candidate items, the optimization model selected a subset of 70 resources, respecting the maximum study time $T_{\max} = 8100$ minutes. The specific index set of selected items from the filtered catalog is presented in Table~\ref{tab:best_solution_indices}.

\begin{table}[h!]
    \centering
    \small
    \begin{equation*}
        \begin{aligned}
            \mathcal{I}^* = \{&
            3, 5, 7, 12, 16, 17, 19, 20, 25, 27, 28, 32, 34, 37, 38, 40, 41, 43, \\
            & 49, 53, 54, 57, 59, 62, 67, 70, 71, 72, 73, 74, 75, 79, 80, 83, 86, 88, \\
            & 90, 91, 99, 106, 111, 115, 117, 125, 127, 130, 131, 134, 141, 142, \\
            & 146, 148, 149, 151, 152, 155, 160, 173, 177, 180, 187, 188, 189, 190, \\
            & 191, 201, 203, 209, 212, 213
            \}.
        \end{aligned}
    \end{equation*}
    \caption{Set of selected resource indices ($\mathcal{I}^*$) corresponding to the best solution found ($f(\mathbf{x}) = 6.075289$).}
    \label{tab:best_solution_indices}
\end{table}

\subsection{Convergence and robustness analysis}

\begin{figure}[h!]
	\centering
	\includegraphics[width=0.85\textwidth]{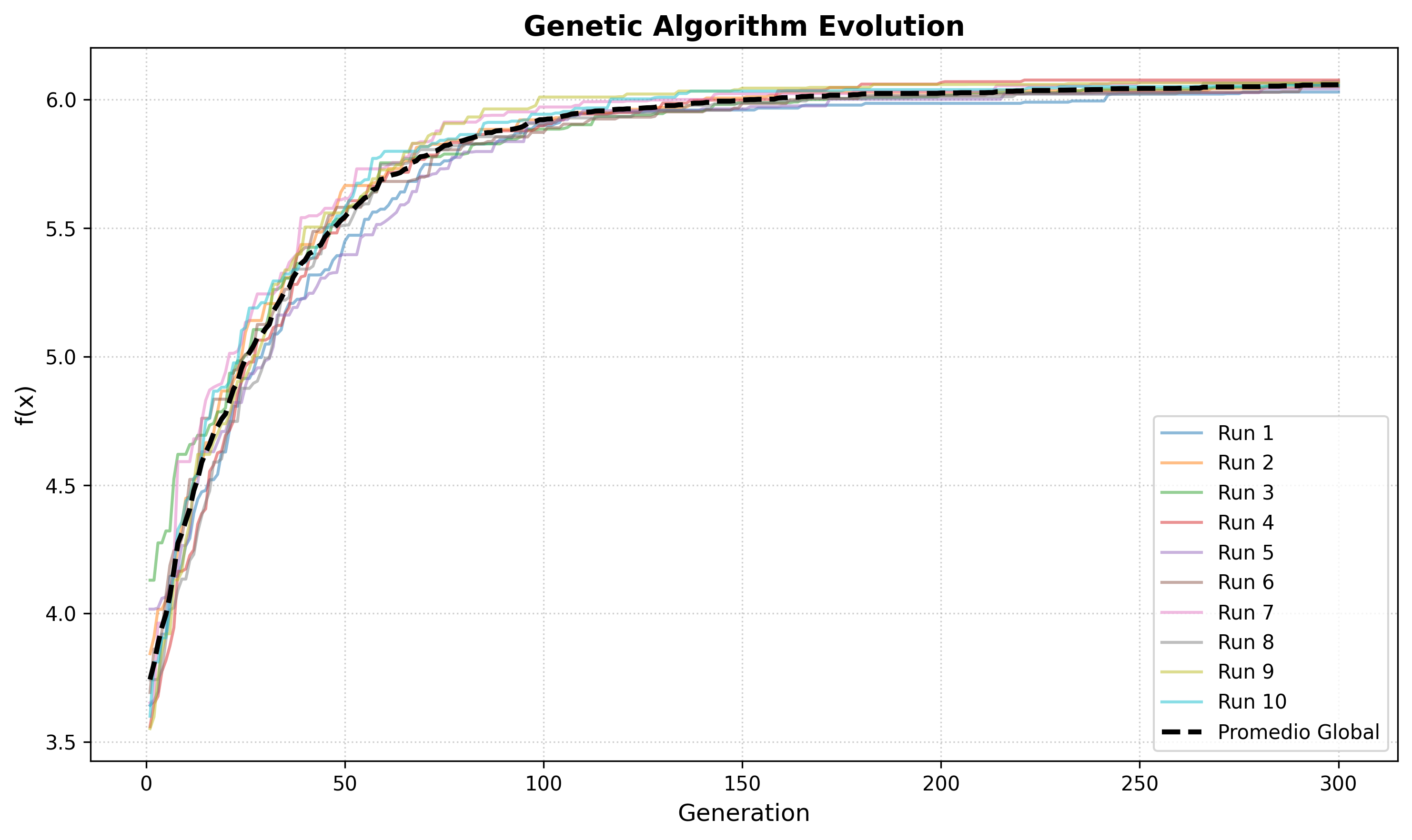}
	\caption{Fitness evolution across 10 independent runs and average convergence trajectory over 300 generations.}
	\label{fig:EvolutionGA}
\end{figure}

Figure~\ref{fig:EvolutionGA} illustrates the evolutionary trajectories across all 10 runs along with the mean convergence curve over the 300 generations.
The algorithm achieves a consistent improvement in fitness throughout the evolutionary process, with the individual runs reaching final fitness values ranging from approximately 6.03 to 6.08 and a mean final fitness of 6.0567. The low standard deviation ($\sigma = 0.0133$) indicates limited variability in the final solutions across the 10 independent runs, suggesting stable and consistent optimization performance despite different random seeds.

\section{Conclusion and future work}
\label{sec:conclusion}

In this paper, we have proposed a metaheuristic-based optimization approach to personalize educational resource selection according to individual student profiles. By incorporating key diagnostic parameters, such as academic level, gap severity, and domain-specific knowledge state, the single-objective formulation maximizes overall pedagogical utility while strictly adhering to time constraint limits. The diagnostic factor controls the penalization of resources whose grade or difficulty level deviates from the student's academic profile, allowing resource selection to be adapted to different educational needs. The experimental evaluation using a Genetic Algorithm implemented via the \texttt{pymoo} framework across 10 independent runs demonstrated consistent convergence behavior, achieving a best fitness value of $f(\mathbf{x}) = 6.0423$.

As future work, we intend to extend this framework to a multi-objective optimization setting tailored for students presenting multiple co-occurring diagnoses. Let $M$ denote the number of distinct diagnoses associated with a given student (e.g., $M=2$ for a student diagnosed with both high abilities and dyslexia). Each diagnosis induces a specific pedagogical criterion modeled through a dedicated objective function. The multi-objective optimization problem is formulated as follows:

\begin{equation}
    \begin{aligned}
        \max_{\mathbf{x}} \quad
        & f_k(\mathbf{x}), \quad k=1,\ldots,M
        \\
        \text{s.t.} \quad
        & g_1(\mathbf{x}) =
        \sum_{i \in \mathcal{R}} \tau_i x_i
        - T_{\max} \leq 0
        \\
        & x_i \in \{0,1\}, \quad \forall i \in \mathcal{R}.
    \end{aligned}
    \label{eq:multi-objective-problem}
\end{equation}

Each objective function $f_k(\mathbf{x})$, for $k=1,\ldots,M$, is defined as an instantiation of Equation~\ref{eq:objective-function} using a diagnosis-specific parameter $s_k \in (0,1]$:

\begin{equation}
    f_k(\mathbf{x})
    =
    \sum_{j \in \mathcal{A}}
    \frac{1}{GE_j}
    \sum_{\substack{i \in \mathcal{R} \\ a_i = j}}
    \frac{s_k}
    {s_k + (c_i - c) + (\ell_i - L_j)}
    x_i.
    \label{eq:objective-functions}
\end{equation}

Unlike a uniform scaling of resource utilities, the diagnosis-specific parameter $s_k$ modifies the relative penalization of academic mismatches. Consequently, different objective functions may induce different preferences over feasible resource allocations, potentially leading to conflicting optimization criteria. When such conflicts arise, the multi-objective formulation enables the identification of Pareto-optimal trade-offs between the pedagogical criteria associated with the student's diagnoses.

\section*{Acknowledgements}
This work has been partially funded by the Erasmus+ KA220-SCH – Cooperation Partnerships in School Education, Ref.: KA220-SCH 0486161A. We also extend our gratitude to the partner middle school and its teachers for their participation in the design and implementation of the exercises, as well as to the mathematics and informatics teams. Our thanks also go to the educational company for its involvement and management in designing the platform used for the mathematical exercises, and to the informatics and psychology professors at the university for their contributions.

\bibliographystyle{unsrt}
\bibliography{references}

\end{document}